\documentclass[conference]{IEEEtran}
\IEEEoverridecommandlockouts
\usepackage{cite}
\usepackage{amsmath,amssymb,amsfonts}
\usepackage{algorithmic}
\usepackage{graphicx}
\usepackage{textcomp}
\usepackage{xcolor}
\usepackage{flushend}
\usepackage{balance}

\usepackage{float}
\usepackage{subfig}
\usepackage{times}
\usepackage{url}
\usepackage{caption}
\def\BibTeX{{\rm B\kern-.05em{\sc i\kern-.025em b}\kern-.08em
    T\kern-.1667em\lower.7ex\hbox{E}\kern-.125emX}}
\begin{document}

\title{YOLO-Based Defect Detection for Metal Sheets\\ 
\thanks{This work was supported in part by the Academia Sinica (AS) under Grant 235g Postdoctoral Scholar Program, in part by the National Science and Technology Council (NSTC) of Taiwan under Grant 113-2926-I-001-502-G.}
}
\author{\IEEEauthorblockN{Po-Heng Chou$^{1}$, Chun-Chi Wang$^{2}$, and Wei-Lung Mao$^{2}$}
\IEEEauthorblockA{$^{1}$Research Center for Information Technology Innovation (CITI), Academia Sinica (AS), Taipei, 11529, Taiwan\\
$^{2}$Department of Electrical Engineering and Graduate School of Engineering Science and Technology,\\ National Yunlin University of Science and Technology (NYUST), Yunlin, 64002, Taiwan\\
E-mails: d00942015@ntu.edu.tw, adsl5253928@gmail.com, wlmao@yuntech.edu.tw}
}

\maketitle
\vspace{-0.5in}
\begin{abstract}
In this paper, we propose a YOLO-based deep learning (DL) model for automatic defect detection to solve the time-consuming and labor-intensive tasks in industrial manufacturing.
In our experiments, the images of metal sheets are used as the dataset for training the YOLO model to detect the defects on the surfaces and in the holes of metal sheets.
However, the lack of metal sheet images significantly degrades the performance of detection accuracy.
To address this issue, the ConSinGAN is used to generate a considerable amount of data.
Four versions of the YOLO model (i.e., YOLOv3, v4, v7, and v9) are combined with the ConSinGAN for data augmentation.
The proposed YOLOv9 model with ConSinGAN outperforms the other YOLO models with an accuracy of 91.3\%, and a detection time of 146 ms.
The proposed YOLOv9 model is integrated into manufacturing hardware and a supervisory control and data acquisition (SCADA) system to establish a practical automated optical inspection (AOI) system.
Additionally, the proposed automated defect detection is easily applied to other components in industrial manufacturing.
\end{abstract}


\section{Introduction}
According to~\cite{r1}, the authors propose a theoretical framework to state a phenomenon that lower population growth promotes investment in adopting automation. 
To meet the requirements of Industry 4.0, it is a trend to apply deep learning (DL) to assist automated production and reduce labor-intensive tasks nowadays.
In the industry, manufacturers of metal sheets are well-known components mass-produced and widely used~\cite{re28_sheets_metal,re29_sheets_metal}.
In the process of automated mass production, the quality management of metal sheets is a critical issue.
However, current quality management of metal sheets relies heavily on human visual inspection.
Especially, the defects of metal sheets, such as scratches on the metal surface, and inconsistent metal holes, are difficult to find.
Thus, a DL-based detection is proposed to inspect the defection of metal sheets.

The You Only Look Once (YOLO)~\cite{re11_YOLOv3,re12_YOLOv4,re_add_5_YOLOv5,re_add_6_YOLOv6, re13_YOLOv7,re_add_7_YOLOv8,YOLOv9} is a DL model for object detection that has shown outstanding performance in terms of accuracy. 
In the training phase, YOLO adopts the feature pyramid network (FPN) framework to enhance learning performance via iterative updating until convergence.
YOLO has been widely used in various real-time vision object detection, such as vehicle detection~\cite{re21}, traffic sign detection~\cite{re16}, and robot vision system~\cite{re19}, etc.
Several studies~\cite{re17,re5,re6,re38,re39,re30_Mao} used YOLO to detect the metal surface defects.
In~\cite{re17}, YOLOv3 was used to detect the defects on hot-rolled steel strips.
In~\cite{re5}, an enhanced YOOLOX algorithm was proposed to detect the defects on the surface of metal sheets.
In~\cite{re6}, YOLOv5, YOLOv7, and region-based convolutional neural networks (RCNN) were used to detect the metal defects and process images from two datasets, including NEV-DET~\cite{re36} and GC10-DET~\cite{re37}. 
In ~\cite{re38}, the YOLOv5 was integrated with the spatial attention module (SAM) and channel attention module (CAM) to detect metal surfaces. 
In~\cite{re39}, YOLOv8 was used to detect the defects of hot rolled strip steel and compared it with a single-shot multi-box detector (SSD) and Faster RCNN.
By reviewing the image detection models~\cite{re_add_1,re_add_2} and comparing the different YOLO models in our previous work~\cite{re30_Mao}, we select YOLOv3~\cite{re11_YOLOv3}, v4~\cite{re12_YOLOv4}, v7~\cite{re13_YOLOv7}, and v9~\cite{YOLOv9}.

However, the datasets of these works are limited to specialized use cases. Even though the public datasets (i.e., NEU-DET~\cite{re36} and GC10-DET~\cite{re37}) are used, it is still not sufficient to be used for pre-training~\cite{re40}. The weakness in metal sheet defect detection is mainly due to the cost of image data collection and labeling. In addition, the dataset is predominated by non-defective samples, which leads to biased data.
To augment the limited amount of data, we use an advanced generative adversarial network (GAN) model~\cite{re21, re41} called ConSinGAN~\cite{re22} to synthesize more image data.
The ConSinGAN can synthesize the image data that involves the defect characteristics based on a single image.
It robustly facilitates the image data augmentation compared to the other GAN models (e.g., deep convolutional GAN (DCGAN)~\cite{re7_DCGAN, re41} and Wasserstein GAN (WGAN)~\cite{re24}, etc).

On the other hand, most of the studies only focus on the accuracy measurements. However, the detection time should be considered as a critical issue in real-time detection applications. In addition, model size should not be increased indefinitely due to computation requirements. It is essential to consider the size of the YOLO model to fit hardware constraints.
Therefore, we integrate the YOLO model into a supervisory control and data acquisition (SCADA) system to control the practical hardware of the automated optical inspection (AOI).

In our previous work~\cite{re30_Mao}, we firstly propose the YOLO-based defect detection with ConSinGAN for dual in-line package (DIP) switch. To show our proposed defect detection without loss of generality, we apply it to detect the defects of metal sheets and augment the insufficient data. The main contributions of this work are as follows:
\begin{itemize}
    \item We design a YOLO-based model to detect the defects for industry components, the paradigm is metal sheets.  
    \item To enhance the performance of the YOLO-based model, we use ConSinGAN to augment image data for pre-training.    
    \item To compare the different YOLO versions, we evaluate the performances in terms of accuracy and detection time.
    \item The proposed YOLOv9 model with ConSinGAN demonstrates an accuracy of 91.3\% and a detection time of 146 ms, outperforming other versions of the YOLO model.
    \item To implement for the user, we develop a  SCADA interface to control the AOI system for practical applications.
\end{itemize}

\section{Automated Defect Detection System}
\label{sec:Automated_System}
\begin{figure}[t]
\centering
{\includegraphics[width=0.35\textwidth]{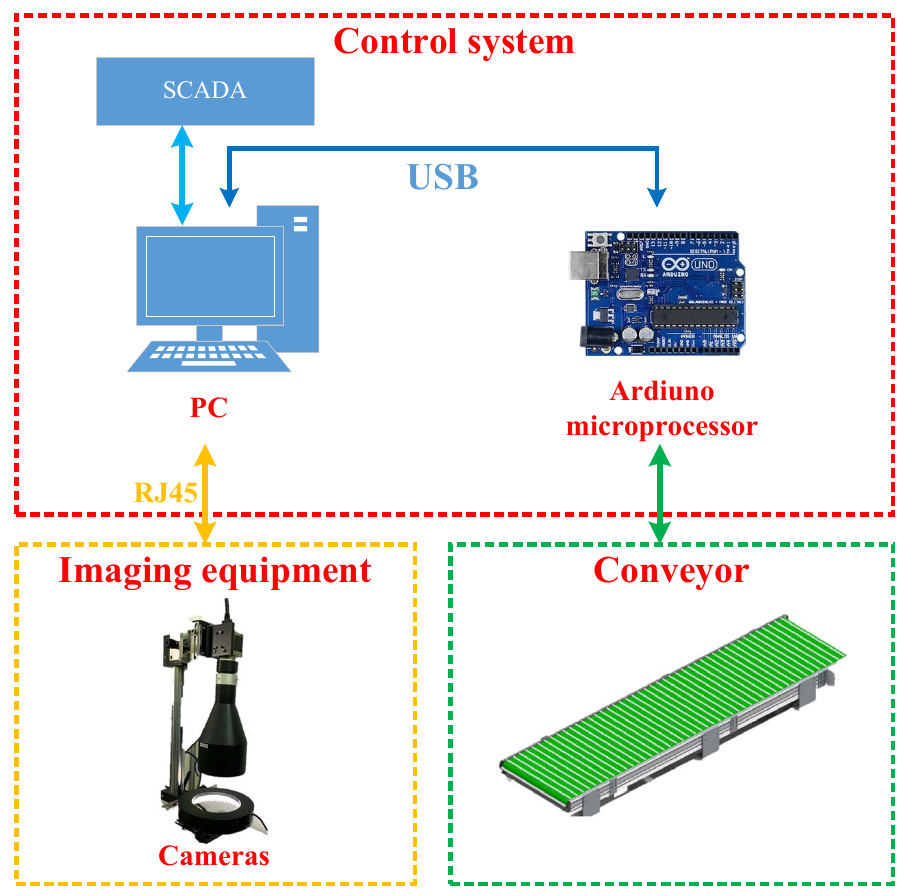}
\caption{The proposed defect detection system architecture.}
\label{fig:1}}
\vspace{-0.2in}
\end{figure}

The proposed system architecture is illustrated in Fig.~\ref{fig:1} and comprises three parts: 
1) Control system: It integrates a personal computer (PC) with an Arduino microprocessor and imaging equipment to establish a SCADA interface for YOLO-based detection. 
2) Imaging equipment: It includes a line-scan camera, a telecentric lens, and various lighting devices, connecting with the PC via Ethernet cable (RJ45) to control image capture actions in the peripheral component interconnect express (PCIe) version. 
3) Conveyor: It is controlled by an Arduino microprocessor that connects to the stepper motor of the conveyor and regulates the speed to enhance image quality.

\subsection{Automated Detection Mechanism}
\begin{figure}[t]
\centering
{\includegraphics[width=0.31\textwidth]{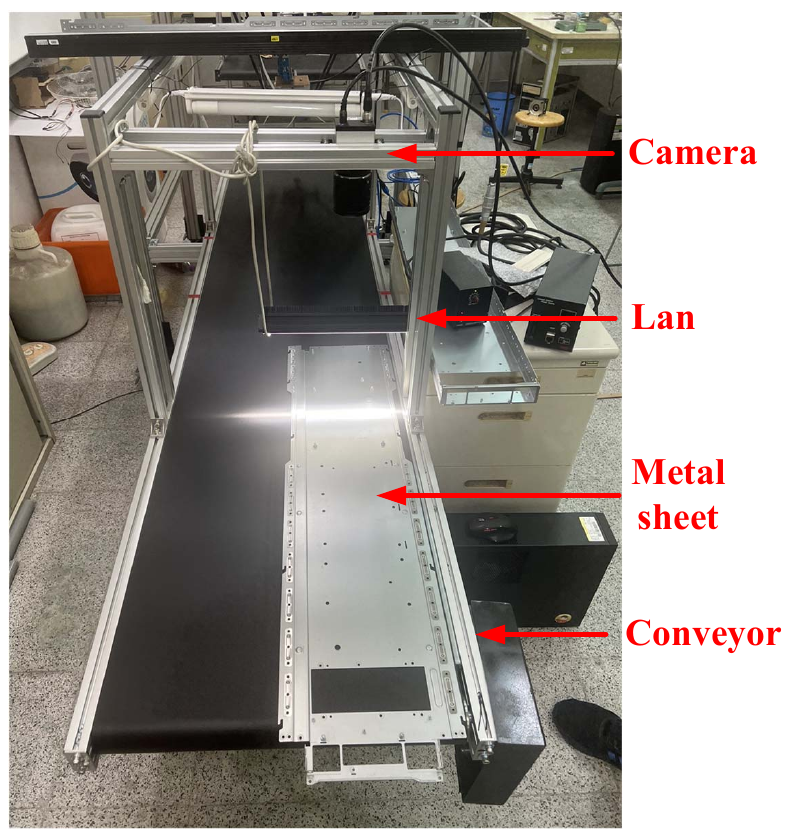}
\caption{The practical defect detection system for metal sheets.}
\label{fig:2}}		
\end{figure}

We establish a practical production line of the automated detection mechanism for workpiece manufacturers.
As shown in Fig.~\ref{fig:2}, the practical detection system includes the line-scan camera and its line illumination technique.
The specifications of the imaging equipment are provided in Table~\ref{tb1:low depth of field}.

\begin{table}[t]
\centering
\caption{Low depth of field camera and lens set specifications.}
\begin{center}
\scalebox{1}{
        \setlength{\tabcolsep}{1mm}
        \begin{tabular}{|c|c|c|c|}\hline
        \multicolumn{2}{|c|}{BASLER-acA4096-30 µm (Camera)} &  \multicolumn{2}{c|}{SPO-200I-4M (Len)}\\
        \hline
        \textbf{Specification}  & \textbf{Parameter} & \textbf{Specification}  & \textbf{Parameter}
        \\\hline
        \textbf{Pixel size}  & 7 µm & \textbf{Depth of field}  & 35mm 
        \\\hline
        \textbf{Image resolution}  & 4096$\times$2 & \textbf{Working distance}  & 240.7mm   
        \\\hline
        \textbf{Magnification}  & 0.14 & &        
        \\\hline
    \end{tabular}}
\label{tb1:low depth of field}
\end{center}
\end{table}

\subsection{The Defects of Metal Sheets}
\begin{figure}[t]
    \centering
    {\includegraphics[width=0.45\textwidth]{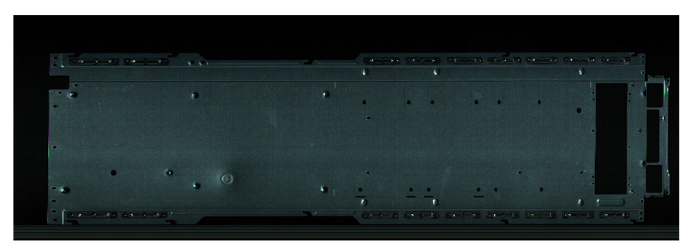}}
    \caption{The practical metal sheets.}
    \label{fig:3}		
\end{figure}

\begin{figure}[t]
    \subfloat[\label{subfig:4(a)}]{
    \includegraphics[width=0.2\textwidth]{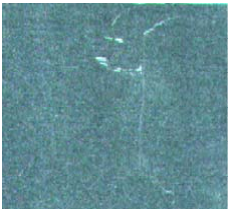}
    }
   \hfill
    \subfloat[\label{subfig:4(b)}]{
    \includegraphics[width=0.2\textwidth]{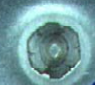}
    }
    \caption{Types of defects. (a) Surface scratches, (b) Irregular holes.}
    \label{fig:4}
    \vspace{-0.2in}
\end{figure}

The metal sheet shown in Fig.~\ref{fig:3} is a semi-finished metal plate that is 2 meters long and 40cm wide. The potential machining errors may lead to the occurrence of scratches on the metal surface, as shown in Fig.~\ref{fig:4} (a), and inconsistencies in the size of holes as shown in Fig.~\ref{fig:4} (b).

\section{YOLO-Based Detection System}
\begin{figure}[t]
    \centering
    {\includegraphics[width=0.49\textwidth]{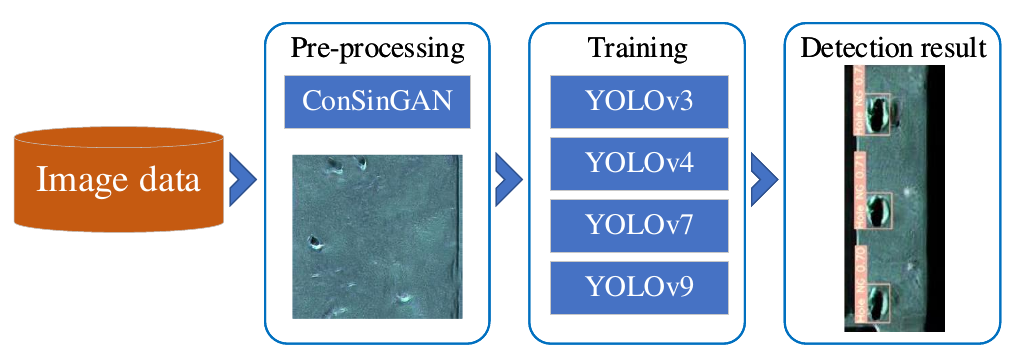}}
    \caption{The flowchart of the proposed defect detection.}
    \label{fig:5}		
    \vspace{-0.2in}
\end{figure}

In this section, we introduce the data pre-processing and augmentation, the adopted YOLO-based models, and performance evaluations.
The workflow of the defect detection system is illustrated in Fig.~\ref{fig:5}.
Since the metal sheet samples are difficult to collect, the number of defect images is inadequate.
To solve the lack of training data, we use the ConSinGan~\cite{re9} based on GAN to augment the image data for the pre-processing. 
Then, we investigate the evolutions among YOLO models v3, v4, v7, and v9.
Finally, the performance metrics, including confusion matrix, precision ($PRE$), recall ($REC$), F1-Score ($F1$), and mean average precision (mAP), are provided.

\subsection{Data Pre-Processing and Augmentation With ConSinGAN}

Due to the lack of image data, the multiple image generative models, such as GAN or DCGAN, may generate the results of overfitting or underfitting, resulting in the loss of the characteristics in the original image.
Therefore, we use the single-image generative model of ConSinGAN to replace the multiple-image generative models, such as GAN or DCGAN. 

The architecture of ConSinGAN includes a multi-stage and multi-resolution approach.
In the first stage, the lowest resolution (e.g., 25 × 25 pixels) is adopted.
In the subsequent stages, the number of layers of the neural network and the resolution of the image are increased. 
At the current training stage, the neural networks in the previous stages are frozen to prevent overfitting and underfitting.
At the current stage $i$, ConSinGAN initializes the discriminator with the weights of the previous stage $i-1$ at all stages, and optimizes the sum of an adversarial loss and a reconstruction loss as follows:
\begin{align}
        \min_{G_{i}} \max_{D_{i}} \mathcal{L}_{adv}(G_{i},D_{i})=\alpha\mathcal L_{rec}(G_{i}),
        \label{eq5}
\end{align}
where $\mathcal{L}_{adv}(G_{i}, D_{i})$ is the adversarial loss~\cite{re24}, the reconstruction loss $L_{rec}(G_{i})$ is used to improve the training stability as follows:
\begin{equation}
        \mathcal{L}_{rec}(G_{i})=||G_{i}(s_{0})-s_{i}||_{2}^{2}.
        \label{eq6}
\end{equation}
In our experiments, $\alpha$ is set to 10.
At the given resolution of stage $i$, $s_{0}$ is input into the generator $G_{i}(s_{0})$. The input $s_{0}$ is a downsampling version of the original image $s_{i}$.

\subsection{YOLO Models v3, v4, v7, and v9}
YOLO is an advanced object detection that was proposed in~\cite{re10} and is well-known due to its capabilities of real-time, efficient detection.
By a single forward in the neural network, YOLO performs object detection, localization, and classification simultaneously~\cite{re_add_1,re_add_2}.
The input image is divided into grids, assigning bounding boxes and class predictions to each grid cell for multi-object detection. 
YOLO locates and identifies objects by predicting bounding boxes and corresponding probabilities of classes. 
The anchor boxes enable the YOLO model to adjust for different object sizes and aspect ratios. 
In addition, the unified loss function considers the localization, confidence, and class predictions, resulting in a comprehensive pre-training. 
However, there are different versions of YOLO.
Based on the comparison of our previous works~\cite{re30_Mao}, we select YOLOv3 ~\cite{re11_YOLOv3}, YOLOv4~\cite{re12_YOLOv4}, YOLOv7~\cite{re13_YOLOv7}, and YOLOv9~\cite{YOLOv9} as the defect detection models. 
The extended features of YOLOv3, v4, v7, and v9 are in the following:

The updates of YOLOv3 are as follows:
\begin{itemize}
    \item It adopts the multiscale feature extraction, FPN, by using the different sizes of convolutional kernels and predicting in three different scales.
    \item The improved classifiers and the integration of Darknet-53, replacing the Darknet-19 network backbone and enhancing the performance of object classification and feature extraction.
    \item It uses binary cross-entropy as the loss function for class predictions to achieve higher accuracy.
\end{itemize}

The updates in YOLOv4 are as follows:
\begin{itemize}
    \item It employs the cross-stage partial network (CSPNet) backbone network to enhance feature extraction and performance. 
    \item The segment anything model (SAM) and path aggregation network (PANet) are introduced to improve the perception and integration of spatial features. 
    \item The multiscale feature extraction~\cite{re25,re26} is adopted to capture targets of different sizes and distances and enhance adaptability to diverse data and scenes. 
    \item It uses complete intersection over union (CIoU)-loss to evaluate the distance between the targeted and the predicted boxes and improve the accuracy.
\end{itemize}

The updates of YOLOv7 are as follows:
\begin{itemize}
    \item A new re-parameterized model, refocusing convolution (RepConv)~\cite{re27}, is used without identity connections to enhance the speed of convergence.
    \item A novel dynamic label assignment strategy is used to enhance feature learning capabilities through hierarchical deep supervision and dynamic label distribution. 
    \item The extended efficient layer aggregation network (E-ELAN), efficiently utilizes parameters and memory usage to enhance the learning capacity without disrupting the original gradient paths.
\end{itemize}

The updates of YOLOv9 are as follows:
\begin{itemize}
    \item The personal growth initiative (PGI) theory involves the auxiliary reversible branch that allows YOLOv9 to generate reliable gradient information during the training phase and pass it to the main branch. 
    \item The deep supervision is improved to reduce loss, and depending on the depth of the model, the size of the integration network is adjusted.
    \item The generalized efficient layer aggregation network (GELAN) is composed of the CSPNet and ELAN networks to balance model size, floating-point operations (FLOPs), and inference speed.
\end{itemize}

\subsection{Performance Metrics}
\label{subsec:Performance evaluation}
We adopt the confusion matrix, $PRE$, $REC$, $F1$, and mAP to measure the detection performances of YOLO models, where $PRE$, $REC$, and $F1$ are shown from Eq.~\eqref{eq3} to \eqref{eq5}.
\begin{equation}
        PRE= \frac{TP}{TP+FP},
        \label{eq3}
\end{equation}
\begin{equation}
        REC= \frac{TP}{TP+FN},
        \label{eq4}
\end{equation}

\begin{equation}
        F1= 2\times\frac{PRE \cdot REC}{PRE+REC},
        \label{eq5}
\end{equation}
where $TP$ (true positive), $TN$ (true negative), $FP$ (false positive), and $FN$ (false negative) represent the cases: the defect and normal metal sheets labeled as such are correctly recognized, the normal metal sheets labeled as defective metal sheets are misclassified, and the defect metal sheets labeled as normal metal sheets are misclassified, respectively.

In addition, mAP is commonly utilized for the assessment of object detection models.
The mAP0.5 is calculated by determining the IoU between the bounding boxes and the ground truth, where mAP0.5 means the threshold of YOLO is set to 0.5. 
In this study, the experiment is conducted on a PC with an Intel 13th Gen I7-13700K CPU (3.40 GHz), NVIDIA RTX4080 graphics card with 16 GB of dedicated memory, running on a 64-bit Windows 10 operating system. 
The training and testing procedures are executed in an Anaconda environment, and the SCADA interface is implemented on Visual Studio 2019.

\section{Experimental Results}
\label{sec:Experimental results}
In the experiments, we compare the performances of YOLOv3, v4, v7, and v9 with/without ConSinGAN. The hyperparameters of the above YOLO models are shown below: Batches = 32, images = 416$\times$416, learning rate = 0.001, and maximum batches = 10000.
The hyperparameter settings for the generative adversarial network ConSinGAN are as follows: Learning rate = 0.1, the number of trained stages = 10. The initial images are obtained by line scanning as mentioned in Sec.~\ref{sec:Automated_System}. 
There are 10 original images of metal sheets. 
The defect characteristics are cropped to dimensions of 512$\times$512 pixels, resulting in 35 images for pre-training. 
The ConSinGAN augments the defect feature images with approximately 20 synthesis images from the original images. 
Our dataset includes 735 images for the pre-training total.

\subsection{Synthesis Images}
\begin{figure}[t]
    \subfloat[\label{subfig:6(a)}]{
    \includegraphics[width=0.2\textwidth]{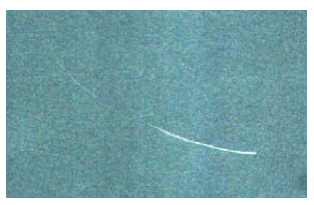}
    }\hspace{+5mm}
    \subfloat[\label{subfig:6(b)}]{
    \includegraphics[width=0.15\textwidth]{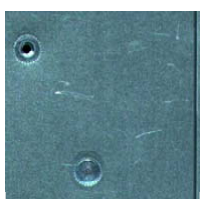}
    }
    \centering
    \caption{Original image. (a) Surface scratches, (b) Hole abnormality.}
    \label{fig:6}

\end{figure}

\begin{figure}[t]
    \subfloat[\label{subfig:7(a)}]{
    \includegraphics[width=0.2\textwidth]{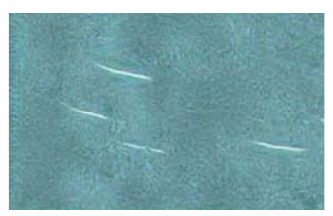}
    }\hspace{+5mm}
    \subfloat[\label{subfig:7(b)}]{
    \includegraphics[width=0.15\textwidth]{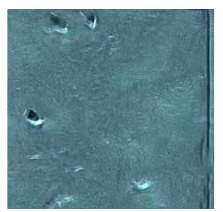}
    }\centering
    \caption{Images generated by ConSinGAN. (a) Surface scratches, (b) Irregular holes.}    \vspace{-0.2in}
    \label{fig:7}
\end{figure}

Initially, we tried to enhance the dataset using GAN and DCGAN. 
Unfortunately, because of the lack of defect image samples, the performance of the YOLO model with GAN or DCGAN is not expected. 
Therefore, we adopt the ConSinGAN, which generates synthesis images only from a single image to augment the dataset effectively, where the original image is displayed as shown in Fig.~\ref{fig:6}, and the synthesis image of the augmentation of ConSinGAN as shown in Fig.~\ref{fig:7}. 
The number of synthesis Images, including surface scratches and irregular holes, is 273 and 462, respectively.


\subsection{Performances of YOLO Models}
\begin{figure}[t]
    \centering
    {\includegraphics[width=0.4\textwidth]{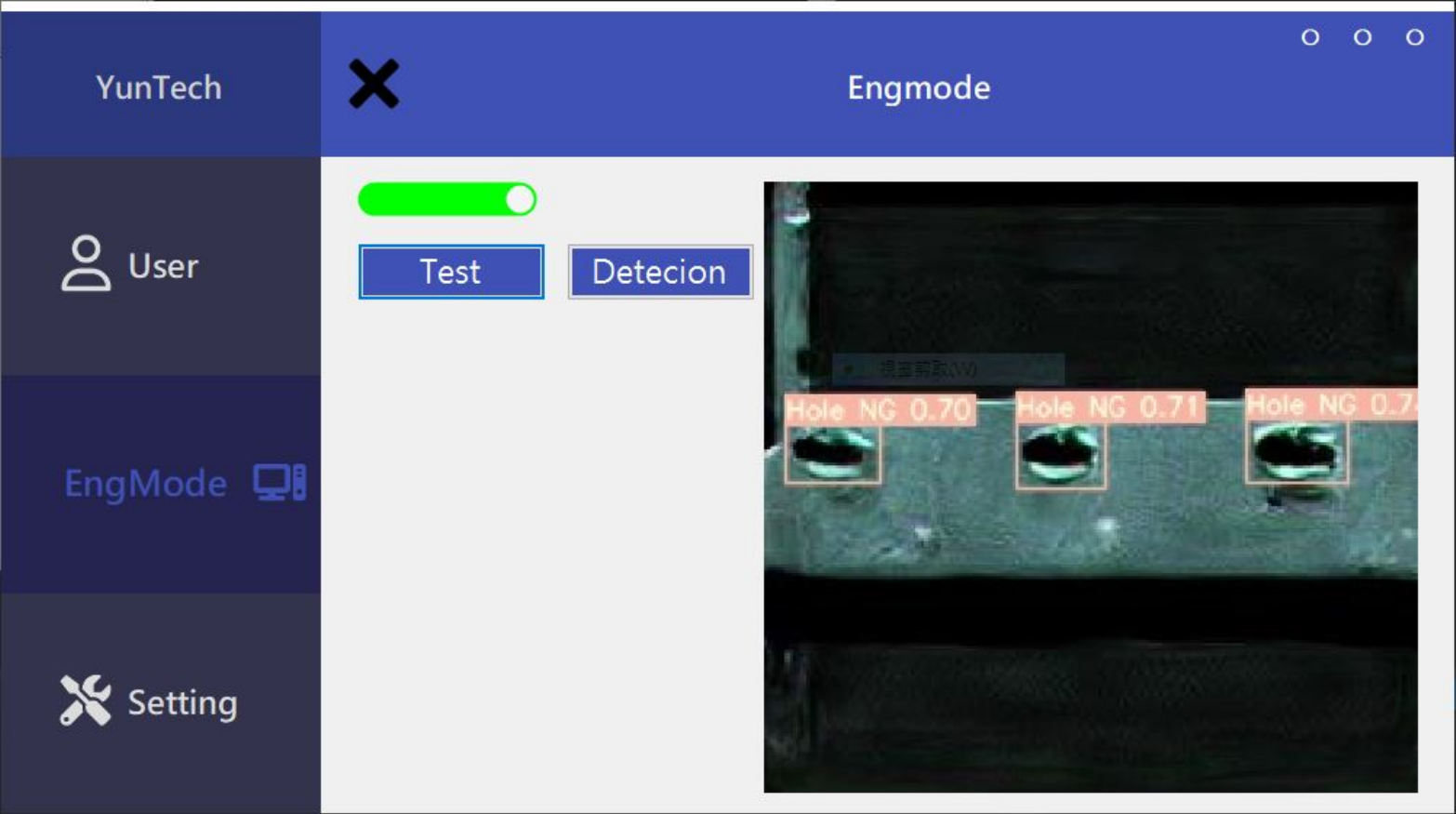}}
    \caption{The SCADA interface.}
    \label{fig:8}
\vspace{-0.2in}
\end{figure}

The dataset is divided into three parts, including the training set, validation set, and testing set.
The training set is conducted on 80\% of the dataset (588 images).
The validation set is adopted on the other 10\% of the dataset (73 images). 
The remaining 10\% (74 images) are used for the testing set. 
Two types of defects can be identifie,d including surface scratches and irregular holes, which are 37.1\% and 62.9\% on our synthesis dataset. 
The performances in terms of the mAP0.5, $PRE$,  $REC$, $F1$, and detection time of the YOLOv3, v4, v7, and v9 models with/without ConSinGAN are presented in Table~\ref{tb3:Yolo model results}.
\begin{table}[t]
\centering
\caption{Performance Comparison of YOLO Models.}
\scalebox{0.8}{
      \begin{tabular}{|c|c|c|c|c|c|c|}
      \hline
       \textbf{Model} & mAP0.5 & \textbf{$PRE$} &\textbf{$REC$} &\textbf{$F1$}  &\textbf{Detection time}\\\hline
       \textbf{YOLOv3} & 70.4\% & 70.1\% &78.5\% &65.4\% &222 ms\\\hline
       \textbf{YOLOv4} & 70.0\% & 76.2\% &74.1\% &68.7\% &224 ms\\\hline
       \textbf{YOLOv7} & 75.3\% & 79.7\% &72.7\% &73.9\% &185 ms\\\hline
        \textbf{YOLOv9} & 79.8\% & 87.5\% &71.6\% &80.1\%&161 ms\\\hline
       \textbf{YOLOv3 with ConSinGAN} &79.4\% & 81.2\% &76.4\% &77.3\% & 221 ms\\\hline
       \textbf{YOLOv4 with ConSinGAN} & 75.1\% & 73.1\% &73.9\% &76.2\% &210 ms\\\hline
       \textbf{YOLOv7 with ConSinGAN} & 88.3\% & 91.8\% &81.5\% &87.9\% &157 ms\\\hline
       \textbf{YOLOv9 with ConSinGAN} & 91.3\% & 98.8\% &85.7\%&91.0\%
       &146 ms\\\hline       
    \end{tabular}}  
\label{tb3:Yolo model results}
\end{table}
As shown in Table~\ref{tb3:Yolo model results}, the YOLO models with ConSinGAN outperform without ConSinGAN in terms of accuracy metrics. 
With ConSinGAN, the performance of YOLOv9 outperforms YOLOv3, v4, and v7 in terms of accuracy metrics.
The average detection time for each model is approximately 210 ms. 
The detection time of YOLOv9 with ConSinGAN is 146 ms, which outperforms other models.
Therefore, the proposed YOLOv9 model is integrated into the practical automated defect detection system. 
In addition, the SCADA interface is accomplished for the users, as shown in Fig.~\ref{fig:8}.

\section{Conclusion}
In this paper, an automated detection system is proposed to detect the defects of metal sheets to reduce the cost of production.
The proposed YOLO detection model is integrated with the manufacturing hardware and SCADA interface as a practical system.
Since there is a lack of defective metallic sheet images, we use ConSinGAN, which is generated based on a single image, to augment the dataset.
In our experiments, we compare the performances of YOLOv3, v4, v7, and v9 with/without ConSinGAN assistance.
The proposed YOLOv9 and ConSinGAN achieve an mAP0.5 of 91.3\% and a detection time of 146 ms, and outperform the other models. 
For future work, we will develop the detection model for similar metallic sheets, such as 3D and curved metal surfaces.
\bibliographystyle{IEEEtran}

\end{document}